\documentclass[10pt,twocolumn,letterpaper]{article}

\usepackage{cvpr}              
\definecolor{cvprblue}{rgb}{0.21,0.49,0.74}
\usepackage[pagebackref,breaklinks,colorlinks,allcolors=cvprblue]{hyperref}
\usepackage{xspace}
\usepackage{amsmath}
\usepackage{multirow}
\usepackage[accsupp]{axessibility}
\usepackage{bm}

\def\paperID{} 
\def\confName{CVPR}
\def\confYear{2026}

\title{Scan Clusters, Not Pixels: A Cluster-Centric Paradigm for Efficient Ultra-high-definition Image Restoration}

\author{Chen Wu$^{1}$\quad Ling Wang$^{2}$\quad Zhuoran Zheng$^{3}$\quad Yuning Cui$^{4}$\\
Zhixiong Yang$^{1}$\quad Xiangyu Chen$^{5}$\quad Yue Zhang$^{6}$\quad Weidong Jiang$^{1}$\quad Jingyuan Xia$^{1,}$\thanks{Corresponding author.}\\
$^{1}$National University of Defense Technology\quad$^{2}$HKUST(GZ)\quad$^{3}$Qilu University of Technology\\
$^{4}$Technical University of Munich\quad$^{5}$TeleAI, China Telecom\quad$^{6}$Beihang University
}
\begin{document}
\maketitle
\footnotetext{This work is supported by NSFC grant.62576350 and 625B2180.}
\begin{abstract}
Ultra-High-Definition (UHD) image restoration is trapped in a scalability crisis: existing models, bound to pixel-wise operations, demand unsustainable computation. While state space models (SSMs) like Mamba promise linear complexity, their pixel-serial scanning remains a fundamental bottleneck for the millions of pixels in UHD content. We ask: must we process every pixel to understand the image? This paper introduces  C$^2$SSM, a visual state space model that breaks this taboo by shifting from pixel-serial to cluster-serial scanning. Our core discovery is that the rich feature distribution of a UHD image can be distilled into a sparse set of semantic centroids via a neural-parameterized mixture model.  C$^2$SSM leverages this to reformulate global modeling into a novel dual-path process: it scans and reasons over a handful of cluster centers, then diffuses the global context back to all pixels through a principled similarity distribution, all while a lightweight modulator preserves fine details. This cluster-centric paradigm achieves a decisive leap in efficiency, slashing computational costs while establishing new state-of-the-art results across five UHD restoration tasks. More than a solution,  C$^2$SSM charts a new course for efficient large-scale vision: scan clusters, not pixels.  The code is available at \url{https://github.com/5chen/C2SSM}.

\end{abstract}    
\section{Introduction}
\label{sec:intro}
With the proliferation of mobile devices and streaming media, Ultra-high-definition (UHD, specifically $3840\times2160$ resolution) imaging has become the dominant paradigm for visual media consumption. However, the pursuit of high-fidelity UHD image restoration (IR) confronts a fundamental and previously unresolved tension: the conflict between the structural redundancy inherent in natural images and the pixel-wise computational primitives employed by contemporary deep models. While State Space Models (SSMs) like Mamba~\cite{mamba} offer linear complexity for long-range dependency modeling, their core operational unit remains the individual pixel. Applying such pixel-serial scanning mechanisms to UHD images (comprising over 8 million pixels) results in prohibitive memory costs and computational load, exceeding the capacity of consumer-grade GPUs and rendering full-resolution modeling impractical.

\begin{figure*}[t!]
    \centering
    \includegraphics[width=.9\linewidth]{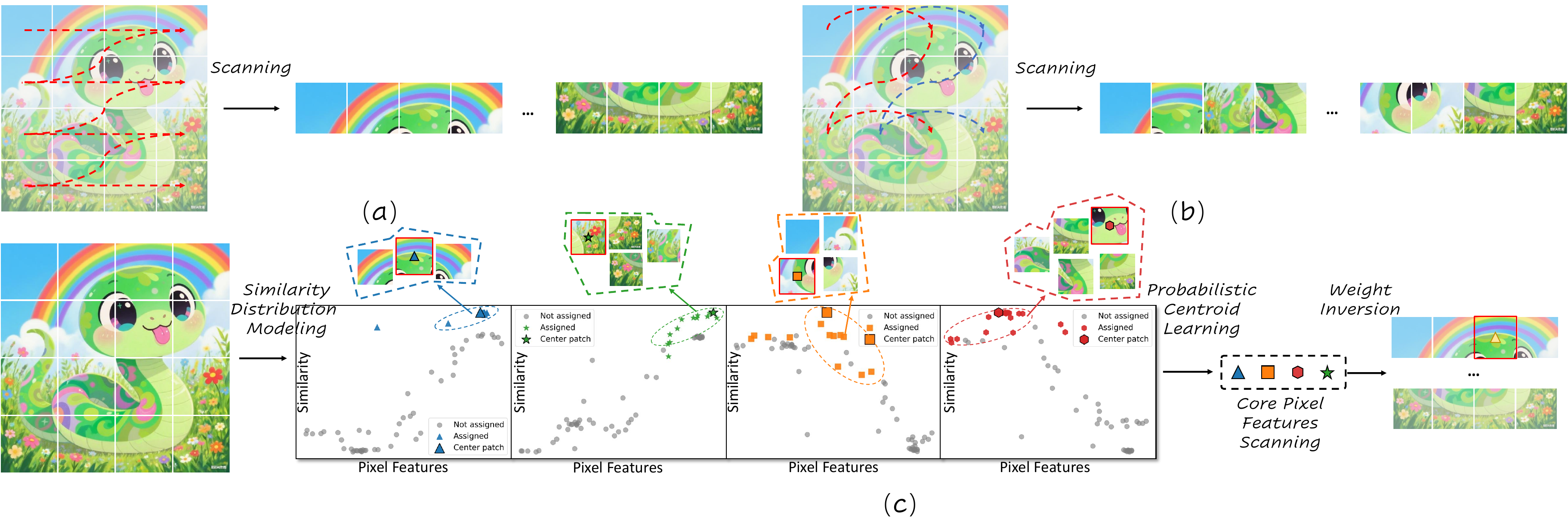}
    \vspace{-10pt}
    \caption{The scanning strategies in existing Mamba-based methods and our proposed method. (a) Vmamba~\cite{Vmamba} employs a Z-shaped scan path that incurs VRAM bottlenecks when processing UHD images due to its full-pixel scanning. (b) EfficientVMamba~\cite{efficientvmamba} reduces scanning costs by omitting sampling steps, this compromises global modeling accuracy. (c) The proposed cluster-centric scanning strategy.}
    \label{fig:diff}
\end{figure*}

Existing attempts to circumvent this bottleneck are fundamentally limited. Multi-scale downsampling methods~\cite{UHDformer,UHDFour,MixNet,UHDHaze,UHDHDR} sacrifice global context and high-frequency details. While SSM-based IR frameworks~\cite{wavemamba,mambairv2,mambair} avoid the quadratic complexity of transformers, they remain bound to pixel- or patch-level scanning, which is intrinsically misaligned with the statistical properties of visual data. These approaches treat pixels as independent entities, failing to capitalize on the underlying low-rank structure and semantic cohesion of image features, thereby incurring substantial and unnecessary computational overhead.  

We posit that the key to efficient UHD restoration lies not in faster pixel processing, but in a paradigm shift from pixel-centric to cluster-centric representation. Natural images are not random collections of pixels; they exhibit strong statistical regularities where features converge into a sparse set of semantically coherent regions. Inspired by this, we introduce C$^2$SSM, a novel visual state space model that reformulates image restoration as a process of neural-parameterized mixture distribution modeling and inference. 

The core of C$^2$SSM is a theoretically grounded dual-path framework: i) The Cluster-Centric Scanning Module (CCSM) explicitly models the feature distribution via a set of learnable cluster centroids. It constructs an n-dimensional similarity distribution to probabilistically associate each pixel with these centroids, effectively reducing the representation space. Global dependencies are then modeled efficiently by applying the SSM only to the sparse centroids, and the learned contextual weights are propagated back to all pixels through a similarity-guided score diffusion process based on the law of total probability. ii) The Spatial-Channel Feature Modulator (SCFM) acts as an information-theoretic compensator. It operates in parallel to preserve high-frequency details that might be attenuated during clustering. The proposed architectural framework transcends conventional engineering optimizations by introducing a novel probabilistic inference model tailored for visual state space systems. This framework facilitates global reasoning by employing a statistically determined sparse graph composed of centroids, while ensuring the preservation of local fidelity through the modulation of complementary features.

The main contributions of this work are threefold:
\begin{itemize}
    \item We introduce C$^2$SSM, the first visual state space model that replaces pixel-level scanning with a cluster-centric probabilistic paradigm. This provides a principled solution to the computational challenges of UHD image restoration.
    \item We design a novel, theoretically principled dual-path framework. The CCSM provides a low-rank approximation for global context modeling via neural-statistical clustering and differentiable weight inversion, while the SCFM ensures local detail preservation.
    \item Through extensive experiments on five UHD restoration tasks, we demonstrate that C$^2$SSM not only achieves state-of-the-art performance but also does so with significantly reduced computational complexity, enabling practical full-resolution restoration on consumer-grade hardware. More importantly, the proposed cluster-centric scanning mechanism offers a new and generalizable direction for efficient large-scale visual computing.
\end{itemize}

\section{Related Work}
\label{sec:related}
\subsection{State Space Model in Image Restoration}
Global receptive fields have proven essential for image restoration tasks~\cite{MixNet,Fourmer,cui2024revitalizing,cuibio}. However, Transformer-based architectures exhibit quadratic computational complexity with respect to input size, resulting in prohibitive computational overhead. Recently, some studies have begun to explore the use of state space models, particularly Mamba, to balance the relationship between efficient computation and direct global receptive fields in restoration tasks. MambaIR~\cite{mambair} applied Visual State Space Models (VSSMs) to image super-resolution, demonstrating competitive performance. MambaIRv2~\cite{mambairv2} subsequently resolved inherent causal modeling limitations. Both FreqMamba~\cite{Freqmamba} and FourierMamba~\cite{Fouriermamba} employ VSSMs for image deraining in the Fourier domain, with FourierMamba introducing enhanced frequency modeling methods. MaIR~\cite{Mair} introduces locality and continuity properties into VSSMs by refining scanning strategies, delivering promising results across multiple image restoration tasks. Despite achieving linear complexity, these methods still encounter VRAM bottlenecks due to the excessive pixel volume of UHD images, rendering them undeployable on consumer-grade GPUs. 

\begin{figure*}[t!]
    \centering
    \includegraphics[width=.95\linewidth]{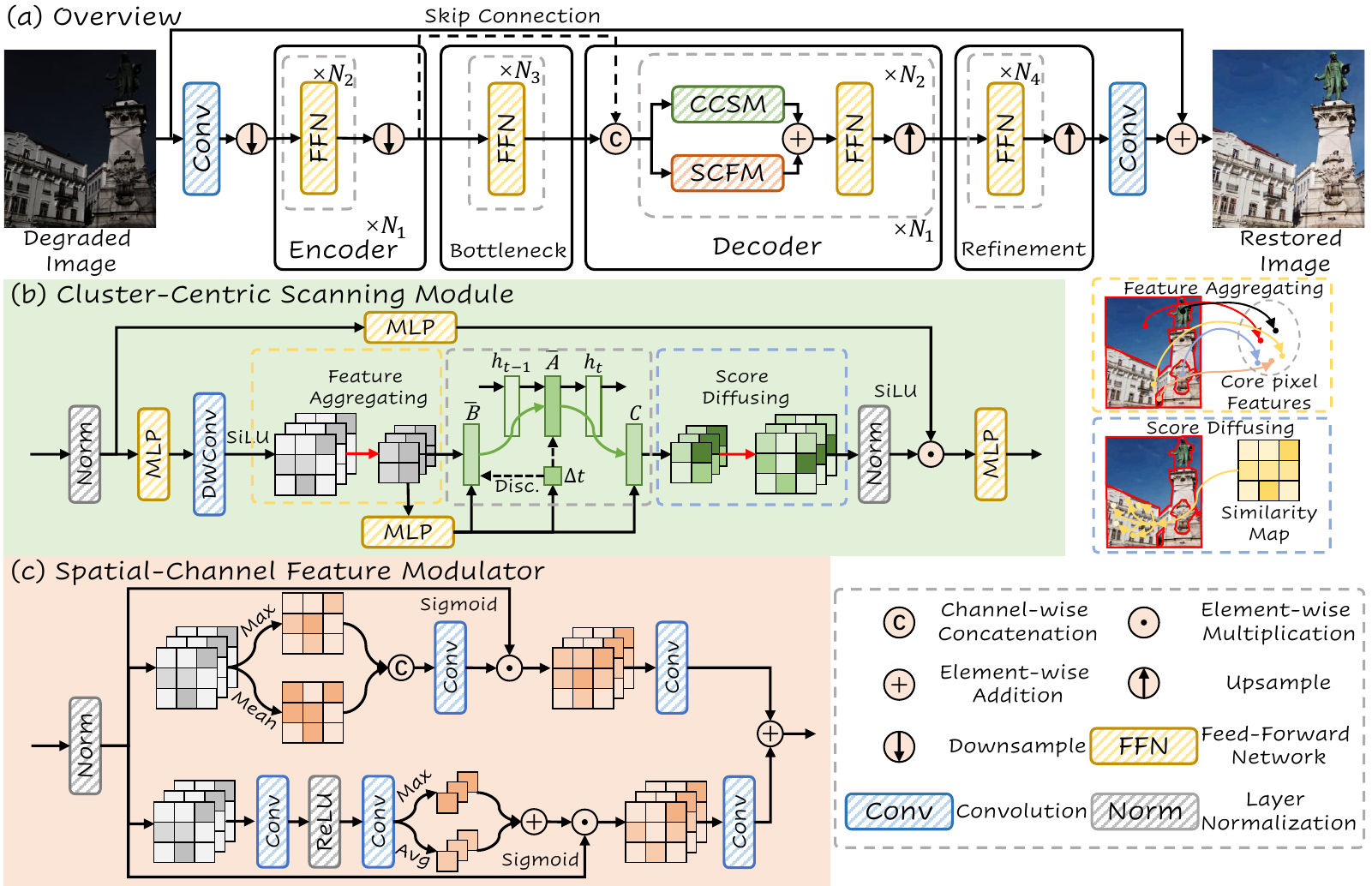}
    \vspace{-10pt}
    \caption{The overview of our proposed C$^2$SSM. C$^2$SSM employs an asymmetric U-Net architecture whose decoder integrates the Cluster-Centric Scanning Module and Spatial-Channel Feature Modulator to achieve spatial-channel global feature coupling.}
    \label{fig:framework}
\end{figure*}

\subsection{UHD Image Restoration}
UHD image restoration has been an emerging topic in recent years. Early approaches predominantly relied on CNNs. For instance, some works extracted local affine coefficients via CNNs and achieved efficient restoration of degraded UHD images through bilateral learning~\cite{UHDHaze,UHDHDR}. DreamUHD~\cite{DreamUHD} and UHD-processor~\cite{UHD-processer} treat the restoration task as a compression-reconstruction task, using retrained VAE encoders instead of common downsampling. They process features in a downscaled latent space and then reconstruct them. Such methods incur high computational costs, and the two-stage training approach can easily lead to error accumulation. With the success of Transformers in computer vision, researchers began exploring their global modeling capacity for restoration tasks. LLFormer~\cite{LLFormer} processes UHD images by splitting them into patches for separate restoration before merging, reducing self-attention overhead but introducing boundary artifacts during reconstruction. UHDformer~\cite{UHDformer} employs a dual-branch strategy, performing global modeling in a highly downsampled low-resolution space. While computationally efficient, this approach discards substantial high-frequency information, compromising restoration quality. The closest work to ours is Wave-Mamba~\cite{wavemamba}, which also leverages a linear-complexity Mamba architecture for long-range modeling. It decomposes images into high/low-frequency components via wavelet transform and models only low-frequency signals. Despite reduced scanning costs, its limited channel capacity (fixed 48 channels throughout the UNet) weakens feature extraction capability, resulting in suboptimal restoration quality. To address this, we propose a novel scanning mechanism for VSSM that revolutionizes full-pixel scanning through cluster-centric point scanning, enabling efficient and effective UHD restoration on consumer-grade GPUs.

\section{Methodology}

To address the prohibitive quadratic complexity of full-pixel scanning in UHD image restoration, we propose a novel probability-driven cluster-centric framework for \(C^2SSM\). The core innovation lies in replacing full-pixel traversal with "centroid learning + global weight inversion"—with the critical cluster assignment and centroid refinement both completed in one-step operations—which achieves an acceptable computational overhead while preserving global context. 

\subsection{Overall Architecture}
As illustrated in Fig.~\ref{fig:framework} (a), the proposed C$^2$SSM adopts an encoder-decoder architecture where degraded images undergo an $N_1$-level restoration pipeline. Each encoder/decoder level contains $N_2$ basic blocks with convolutional sampling layers (down/up-sampling). Following prior works~\cite{fftformer,AST}, an asymmetric design is implemented: the encoder comprises only FFNs to reduce computational load, while the decoder, inspired by MetaFormer~\cite{metaformer}, integrates our CCSM and the SCFM alongside FFNs. A bottleneck layer between the encoder and the decoder allows deep feature extraction, with skip connections via $1\times1$ convolutions incorporating encoder features into decoder layers. A feature refinement stage post-decoder enhances learned representations, and the final restored image is obtained by adding the learned residual to the degraded input.

\subsection{Cluster-Centric Scanning Module}
Visual images naturally contain a high degree of semantic redundancy due to the tendency of spatially adjacent areas to share converging feature weight patterns. In addressing this, CCSM employs feature aggregating to concentrate on contextually aggregated significant pixels, thereby substantially decreasing the computational burden associated with global scanning models. Following this, score diffusing enriches the data available for non-essential pixels, facilitating the regional-level reconstruction of entire areas from sparse center points. The CCSM architecture is depicted in Fig.~\ref{fig:framework} (b). For a layer normalized input feature maps $\boldsymbol{F}_{in}$, the calculation of CCSM is updated to:
\begin{align}
\boldsymbol{F}_{d}&=\operatorname{SiLU}(\operatorname{DWConv}(\operatorname{MLP}(\boldsymbol{F}_{in}))),\\
\boldsymbol{F}_{f}&=\operatorname{Norm}(\operatorname{SD}(\operatorname{S6}(\operatorname{FA}(\boldsymbol{F}_{d})))),\\
\boldsymbol{F}_{out}&=\boldsymbol{F}_{f}\cdot \operatorname{SiLU}(\operatorname{MLP}(\boldsymbol{F}_{in})),
\end{align}
where $\operatorname{SiLU}(\cdot)$ is the SiLU activation functions. $\operatorname{Norm}(\cdot)$ denotes the normalization layer. $\operatorname{FA}(\cdot)$ and $\operatorname{SD}(\cdot)$ are designed feature aggregating and score diffusing, respectively. $\operatorname{S6}$ represents the selective scanning mechanism proposed by Mamba~\cite{mamba}.
\subsubsection{Feature Aggregating}

This stage aims to learn a set of effective, semantically representative centroids from UHD image features, avoiding the inefficiency of random or space-constrained clustering. The key is to model the similarity between pixels and initial centroids as a probabilistic distribution, enabling one-step cross-spatial pixel assignment and one-step adaptive centroid refinement without any iterative processes, ensuring computational efficiency.

\noindent\textbf{Initial Centroid Initialization}:
Given the layer-normalized feature tensor \(F \in \mathbb{R}^{C \times H \times W}\) (output from the encoder), we first select $n$ initial centroids \(\{c_1, c_2, ..., c_n\}\) where \(c_k \in \mathbb{R}^{C \times 1 \times 1}\). 
The initialization follows a uniform sampling strategy across the feature space: we randomly select $n$ pixel positions and calculate their $k$-nearest neighbor values to enhance local bias. 
This ensures initial centroids cover diverse feature patterns of the UHD image.

\noindent\textbf{$n$-Dimensional Similarity Distribution Modeling}:
For each initial centroid \(c_k\), we compute the cosine similarity between every pixel feature \(F_{:,i,j}\) (flattened as \(f_p\) to construct a 1-dimensional similarity distribution \(D_k\) with \(c_k\)). 
Collectively, the $n$ centroids form an $n$-dimensional similarity distribution \(\mathcal{D} = \{D_1, D_2, ..., D_n\}\), where each dimension \(D_k\) is defined as a probability density function (PDF):
\begin{equation}
    p_k(f_p) = \frac{sim(f_p, c_k)}{\sum_{p \in \Omega} sim(f_p, c_k)}, 
    \label{eq:pdf}
\end{equation}
where \(\Omega\) denotes all pixels in the UHD image, and \(sim(\cdot, \cdot)\) is the cosine similarity:
\begin{equation}
    sim(f_p, c_k) = \frac{f_p^T \cdot c_k}{\|f_p\| \cdot \|c_k\|}. \label{eq: 1}
\end{equation}
For \(D_k\), the horizontal axis represents the feature value of pixels (projected to 1D via PCA for interpretability), and the vertical axis represents the normalized similarity (\ie, the probability that the pixel belongs to the cluster dominated by \(c_k\)). This $n$-dimensional distribution effectively models the semantic correlation between each pixel and the $n$ centroids, transforming pairwise similarity into a probabilistic association.

\noindent\textbf{Centroid Refinement via Learnable Function}: 
For each initial centroid \(c_k\), precomputed similarity distribution \(p_k(f_p)\) between the centroid and each pixel feature \(f_p\), the refined centroid \(\hat{c}_k\) is obtained through adaptive feature aggregation guided by a learnable gating mechanism. The calculation incorporates two learnable parameters that adjust the sensitivity of similarity-based pixel selection to adapt to diverse feature patterns across different clusters and datasets. Similar to the $qkv$ mechanism in self-attention~\cite{vaswani2017attention}, we do not directly compute it; instead, we first use an MLP to map $c_k$ and $f_p$ to $v_k$ and $\hat{f}_p$, respectively. The refined centroid is formulated as
\begin{equation}
    \hat{c}_k = \frac{1}{N_k} \left( v_k + \sum_{p \in \Omega} \delta(\alpha \cdot p_k(f_p) + \beta) \cdot \hat{f}_p \right),\label{centroid}
\end{equation}
where the gating function \(\delta(\cdot)\) employs a smooth activation to softly select pixels with meaningful similarity to the initial centroid, balancing selectivity and gradient flow during training. The normalization factor \(N_k\) is derived from the sum of activated gating values plus one, ensuring the initial centroid’s contribution is retained while scaling the aggregated pixel features to maintain numerical stability. This factor is calculated as
\begin{equation}
    N_k = 1 + \sum_{p \in \Omega} \delta(\alpha \cdot p_k(f_p) + \beta).
\end{equation}
The learnable scaling parameter $\alpha$ modulates the sharpness of similarity-based selection, increasing to enforce stricter relevance thresholds in edge-dominated regions and decreasing to include more diverse features in texture-rich areas. The learnable bias parameter $\beta$ shifts the activation threshold, adapting to the overall similarity distribution of each cluster to avoid over-pruning or under-selection of relevant pixels. This gating mechanism inherently prunes pixels with insignificant similarity to the centroid, reducing the number of effective computations while preserving semantic relevance.

\subsubsection{Score Diffusing}
This stage leverages Mamba’s strengths in long-range dependency modeling but only applies it to the $n$ refined centroids (instead of all pixels), then inverts the global pixel weights based on the $n$-dimensional similarity distribution. The process mimics Transformer’s attention mechanism but avoids full-pixel pairwise computation.

\noindent\textbf{Mamba-Based Centroid Weight Estimation}: We feed the refined centroids \(\hat{C} = [\hat{c}_1, \hat{c}_2, ..., \hat{c}_n] \in \mathbb{R}^{C \times n}\) into Mamba’s selective scanning module (S6 block) to learn their precise global weights. Mamba’s state-space modeling efficiently captures long-range dependencies between centroids, outputting a set of centroid-specific weights \(W = [w_1, w_2, ..., w_n] \in \mathbb{R}^{C \times n}\), where \(w_k\) denotes the global context weight of centroid \(\hat{c}_k\):
\begin{equation}
    W = S6(\hat{C}; \theta_{mamba}). \label{eq: 4}
\end{equation}
Here, \(\theta_{mamba}\) are the learnable parameters of the Mamba module. The complexity of this step is \(O(C \cdot n^2)\), which is negligible compared to \(O(C \cdot H^2W^2)\) for full-pixel scanning (since \(n \ll HW\), although there are dimensional transformations in the network that result in an unequal number of channels, all channels are still of the same order of magnitude.)



\noindent\textbf{Weight Inversion via Similarity Distribution}: We formalize the assignment probability \(\alpha_{p,k}\) of pixel $p$ to cluster $k$ as the posterior probability derived from the $n$-dimensional similarity distribution \(\mathcal{D}\). Unlike independent parameterization, \(\alpha_{p,k}\) is directly normalized from the similarity distribution \(p_k(f_p)\) (from Eq.~\ref{eq:pdf}) to retain probabilistic consistency:
\begin{equation}
\alpha_{p, k}=\frac{\exp\left(\alpha \cdot p_{k}\left(f_{p}\right)+\beta\right)}{\sum_{k'=1}^{n} \exp\left(\alpha \cdot p_{k'}\left(f_{p}\right)+\beta\right)}. \label{eq: 5}
\end{equation}
Here, \(\alpha_{p, k}\) quantifies the probability that pixel p belongs to the cluster dominated by centroid \(\hat{c}_{k}\). We adopt softmax normalization to strictly satisfy the probability axiom \(\sum_{k=1}^{n} \alpha_{p, k}=1\), where \(\alpha\) and \(\beta\) are learnable parameters modulating the sharpness of the distribution. 
This definition directly links the weight inversion to the earlier similarity distribution modeling, forming a closed probabilistic loop.
Based on the law of total probability, the global weight \(w_p\) of pixel $p$ is the expected value of the centroids’ weights \(W = [w_1, w_2, ..., w_n]\) (from Eq.~\eqref{eq: 4}), conditioned on the pixel’s similarity distribution \(\mathcal{D}\). The inversion formula is thus:
\begin{equation}
w_p = \mathbb{E}_{k \sim \mathcal{D}(p)} [w_k] = \sum_{k=1}^n \alpha_{p,k} \cdot w_k \label{eq: 6}
\end{equation}
where \(w_p\) denotes the global weight of pixel $p$, and the expectation \(\mathbb{E}_{k \sim \mathcal{D}(p)} [w_k]\) explicitly emphasizes that the weight is computed based on the probability distribution of the pixel across the $n$ clusters.

\subsection{Spatial-Channel Feature Modulator}
To address potential high-frequency detail loss caused by centroid-based aggregation, SCFM operates in parallel with the weight inversion stage. It employs dual-branch attention (spatial + channel) to maximize mutual information between input and output features \cite{Cbam}:
\begin{align}
\boldsymbol{W}_s&=\delta(\operatorname{Conv}([\operatorname{Max}(\boldsymbol{F}_{in});\operatorname{Mean}(\boldsymbol{F}_{in})])),\\
\boldsymbol{F}_{d}&=\operatorname{Conv}(\operatorname{ReLU}(\operatorname{Conv}(\boldsymbol{F}_{in})),\\
\boldsymbol{W}_c&=\delta(\operatorname{Max}(\boldsymbol{F}_{d})+\operatorname{Avg}(\boldsymbol{F}_{d})),\\
\boldsymbol{F}_{out}&=\operatorname{Conv}(\boldsymbol{W}_s \cdot \boldsymbol{F}_{in}) + \operatorname{Conv}(\boldsymbol{W}_c \cdot \boldsymbol{F}_{in}),
\end{align}
where $\operatorname{Max}(\cdot)$ refers to the maximum value operation, while $\operatorname{Mean}(\cdot)/\operatorname{Avg}(\cdot)$ denotes the average value operation. $[\ ;\ ]$ is the concatenation operation. $\operatorname{ReLU}(\cdot)$ represents ReLU activation function.
\begin{figure*}[t!]
    \centering
    \includegraphics[width=.85\linewidth]{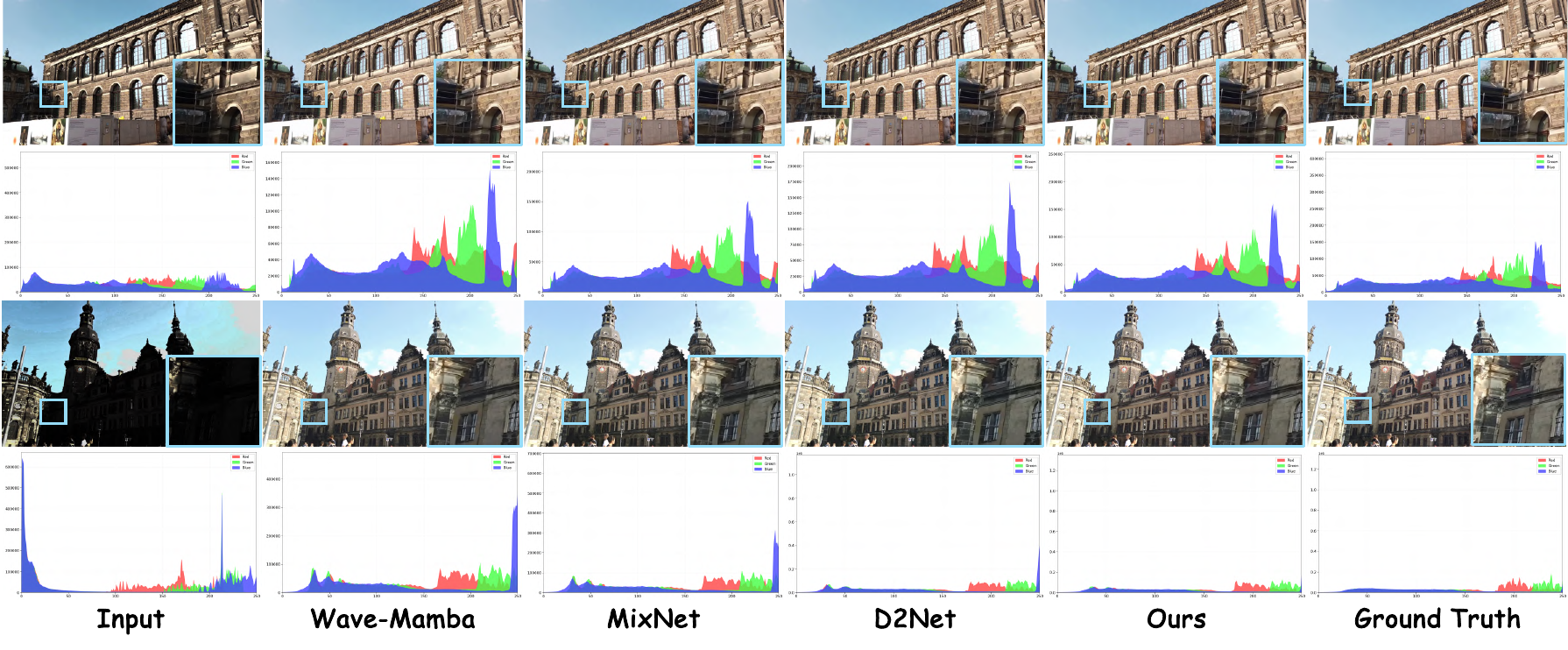}
    \vspace{-10pt}
    \caption{Visual quality comparisons on UHD-LOL4K dataset~\cite{LLFormer}. The last row shows the color histogram of the image.}
    \label{fig:lol4k}
    \vspace{-8pt}
\end{figure*}

\section{Experiments}
\subsection{Experimental Settings}
\noindent\textbf{Implementation Details}: 
Our experiments are conducted using PyTorch on a setup of 4 NVIDIA A800 GPUs. To optimize the network, we employ the AdamW optimizer with a initial learning rate $5 \times 10^{-4}$ and a cosine annealing strategy is used for the decay of the learning rate. We randomly crop the full-resolution 4K image to a resolution of $768\times768$ as the input and the batch size is set to 16. For all UHD restoration tasks, we perform 150K iterations. To augment the training data, random horizontal and vertical flips are applied to the input images. Our method consists of an encoder-decoder with $N_1=3$ levels, where both the encoder and decoder share the same block structure: $N_2 = [2, 4, 4]$. The bottleneck and refinement stages each contain $N_3=N_5=4$ blocks, with a basic embedding dimension of 32. To optimize the weights and biases of the network, we utilize the L1 loss and the FFT loss in the RGB color space as the basic reconstruction loss.


\noindent\textbf{Evaluation}: 
We utilize PSNR~\cite{PSNR} and SSIM~\cite{SSIM} to assess images with ground truth, while NIQE~\cite{NIQE} and PIQE~\cite{PIQE} are used for images without it. Elevated PSNR/SSIM values signify enhanced performance, and diminished NIQE/PIQE scores reflect improved quality. Moreover, we compared model parameters across all techniques.
\begin{table}[t!]
\centering
\caption{Comparison of quantitative results on UHD-LOL4K dataset~\cite{LLFormer}.}
\vspace{-10pt}
\resizebox{\linewidth}{!}{
\begin{tabular}{c|c|c|cc|c}
\toprule[0.5mm]
Methods    & Type                                          & Venue    & PSNR  & SSIM  & Param  \\ \hline
Z\_DCE++~\cite{ZDCE}   & \multicolumn{1}{c|}{\multirow{3}{*}{non-UHD}} & TPAMI'21 & 15.58 & 0.934 & 79.42K\\
Uformer~\cite{Uformer}    & \multicolumn{1}{c|}{}                         & CVPR'22  & 29.98 & 0.980 & 20.63M\\
Restormer~\cite{Restormer}  & \multicolumn{1}{c|}{}                         & CVPR'22  & 36.90 & 0.988 & 26.11M\\ \hline
NSEN~\cite{NSEN}       & \multicolumn{1}{c|}{\multirow{8}{*}{UHD}}     & MM'23    & 29.49 & 0.980 & 2.67M\\
UHDFour~\cite{UHDFour}    & \multicolumn{1}{c|}{}                         & ICLR'23  & 36.12 & \underline{0.990} & 17.54M\\
LLFormer~\cite{LLFormer}   & \multicolumn{1}{c|}{}                         & AAAI'23  & 37.33 & 0.988 & 24.52M \\
UHDformer~\cite{UHDformer}  & \multicolumn{1}{c|}{}                         & AAAI'24  & 36.28 & 0.989 & 0.34M \\
Wave-Mamba~\cite{wavemamba} & \multicolumn{1}{c|}{}                         & MM'24    & 37.43 & \underline{0.990} & 1.25M \\
MixNet~\cite{MixNet}     & \multicolumn{1}{c|}{}                         & NeuroC'24 & \underline{39.22} & \textbf{0.992} & 7.77M \\
D2Net~\cite{D2Net}      & \multicolumn{1}{c|}{}                         & WACV'25  & 37.73 & \textbf{0.992} & 5.22M \\
ours       & \multicolumn{1}{c|}{}                         & -        & \textbf{39.61} & \textbf{0.992} & 2.71M\\
\bottomrule[0.5mm]
\end{tabular}}
\label{tab:UHDLOL}
\vspace{-0.4cm}
\end{table}
\begin{table}[t!]
\caption{Comparison of quantitative results on UHD-LL dataset~\cite{UHDFour}.}
\vspace{-10pt}
\label{tab:UHDLL}
\resizebox{\linewidth}{!}{
\begin{tabular}{c|c|c|cc|c}
\toprule[0.5mm]
Methods   & Type                                          & Venue    & PSNR   & SSIM & Param \\ \hline
Z\_DCE++~\cite{ZDCE}  & \multicolumn{1}{c|}{\multirow{3}{*}{non-UHD}} & TPAMI'21 & 16.41 & 0.630 & 10.56K\\
Uformer~\cite{Uformer}   & \multicolumn{1}{c|}{}                         & CVPR'22  & 19.28 & 0.849 & 20.63M\\
Restormer~\cite{Restormer} & \multicolumn{1}{c|}{}                         & CVPR'22  & 22.25 & 0.871 & 26.11M\\ \hline
UHDFour~\cite{UHDFour}      & \multicolumn{1}{c|}{\multirow{8}{*}{UHD}}  & ICLR'23    & 26.22 & 0.900 & 17.54M\\
LLFormer~\cite{LLFormer}   & \multicolumn{1}{c|}{}                         & AAAI'23  & 22.79 & 0.853 & 13.15M\\
UHDformer~\cite{UHDformer}  & \multicolumn{1}{c|}{}                         & AAAI'24  & 27.11  & 0.927 & 0.34M \\
Wave-Mamba~\cite{wavemamba} & \multicolumn{1}{c|}{}                         & MM'24  & 27.35  & 0.913  & 1.26M \\
UHDDIP~\cite{UHDDIP} & \multicolumn{1}{c|}{}                         & TCSVT'25  & 26.74  & 0.928  & 0.81M \\
MixNet~\cite{MixNet}    & \multicolumn{1}{c|}{}                         & NeuralC’25  & \underline{27.54} & 0.862 & 7.77M\\
UHD-processer~\cite{UHD-processer} & \multicolumn{1}{c|}{}                         & CVPR’25 & 27.22 & \textbf{0.929} & 1.6M\\
Ours    & \multicolumn{1}{c|}{}                         & -        & \textbf{27.63} & \textbf{0.931} & 2.71M\\
\bottomrule[0.5mm]
\end{tabular}}
\end{table}
\subsection{Comparisons with the State-of-the-art Methods}

\noindent\textbf{Low-light Image Enhancement Results}: 
For the task of enhancing UHD images in low-light conditions~\cite{LLFormer,UHDFour}, we evaluate C$^2$SSM against techniques such as Z\_DCE++~\cite{ZDCE}, Uformer~\cite{Uformer}, Restormer~\cite{Restormer}, NSEN~\cite{NSEN}, UHDFour~\cite{UHDFour}, LLFormer~\cite{LLFormer}, UHDformer~\cite{UHDformer}, Wave-Mamba~\cite{wavemamba}, MixNet~\cite{MixNet}, D2Net~\cite{D2Net}, UHDDIP~\cite{UHDDIP} and UHD-processer~\cite{UHD-processer}. As evidenced in Tabs.~\ref{tab:UHDLOL} and~\ref{tab:UHDLL}, our method outoperforms the current SOTA MixNet by 0.39 dB and 0.19 dB in PSNR on synthetic and real-world datasets, respectively. Compared to the Mamba-based method Wave-Mamba, it achieves a significant 2.18 dB improvement. Visual comparisons in Fig.~\ref{fig:lol4k} demonstrate that our approach delivers superior color correction, reconstructing images with enhanced visual fidelity and structural clarity.

\begin{table}[t!]
\centering
\caption{Comparison of quantitative results on 4K-Rain13k dataset~\cite{UDR-Mixer}.}
\vspace{-10pt}
\resizebox{\linewidth}{!}{
\begin{tabular}{c|c|c|cc|c}
\toprule[0.5mm]
Methods      & Type                                          & Venue    & PSNR  & SSIM  & Param \\ \hline
RCDNet~\cite{RCDNet}       & \multicolumn{1}{c|}{\multirow{8}{*}{non-UHD}} & CVPR’20  & 30.83 & 0.921 & 3.17M\\
SPDNet~\cite{SPDNet}       & \multicolumn{1}{c|}{}                         & ICCV'21  & 31.81 & 0.922 & 3.04M\\
IDT~\cite{IDT}          & \multicolumn{1}{c|}{}                         & TPAMI'22 & 32.91 & 0.948 &  16.41M\\
Restormer~\cite{Restormer}    & \multicolumn{1}{c|}{}                         & CVPR'22  & 33.02 & 0.934 & 26.12M\\
DRSformer~\cite{DRSformer}    & \multicolumn{1}{c|}{}                         & CVPR'23  & 32.96 & 0.933 & 33.65M\\
UDR-S2Former~\cite{UDR-S2Former} & \multicolumn{1}{c|}{}                         & ICCV'23  & 33.36 & 0.946 & 8.53M\\
NeRD-Rain~\cite{NeRD-Rain}    & \multicolumn{1}{c|}{}                         & CVPR'24  & 33.63 & 0.935 & 22.9M\\
MambaIRv2~\cite{mambairv2}    & \multicolumn{1}{c|}{}                         & CVPR'25  & 33.17 & 0.939 & 12.7M\\ \hline
UDR-Mixer~\cite{UDR-Mixer}    & \multicolumn{1}{c|}{\multirow{3}{*}{UHD}}     & TMM'25 & 34.30 & 0.951 & 4.90M\\
ERR~\cite{ERR}          & \multicolumn{1}{c|}{}                         & CVPR'25  & \underline{34.48} & \underline{0.952} & 1.13M\\
Ours         & \multicolumn{1}{c|}{}                         & -        & \textbf{35.13} & \textbf{0.956} & 2.71M\\
\bottomrule[0.5mm]
\end{tabular}}
\label{tab:4KRain}
\end{table}

\begin{figure*}[t!]
    \centering
    \includegraphics[width=.85\linewidth]{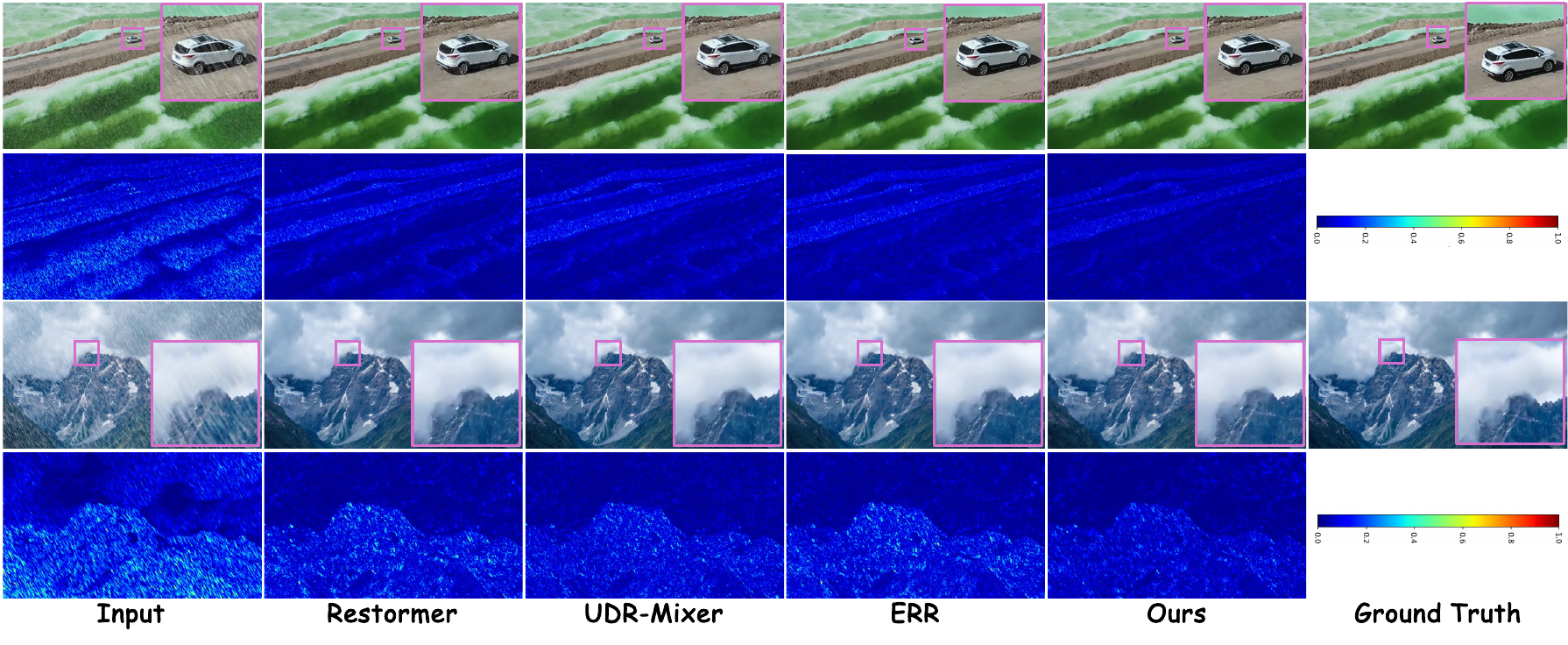}
    \vspace{-10pt}
    \caption{Visual quality comparisons on 4K-Rain13k dataset~\cite{UDR-Mixer}. The last row shows the error map of the image.}
    \label{fig:rain}
    \vspace{-8pt}
\end{figure*}
\begin{table*}[t!]
\centering
\caption{Comparison of quantitative results on 4K-RealRain dataset~\cite{UDR-Mixer}.}
\vspace{-10pt}
\resizebox{.9\linewidth}{!}{
\begin{tabular}{c|ccccccccc}
\toprule[0.35mm]
Methods & RCDNet~\cite{RCDNet} & SPDNet~\cite{SPDNet} & IDT~\cite{IDT}    & Restormer~\cite{Restormer} & DRSformer~\cite{DRSformer} & NeRD-Rain~\cite{NeRD-Rain} & MambaIRv2~\cite{mambairv2} & ERR~\cite{ERR} & Ours  \\ \hline
NIQE    & 9.997  & 9.917  & 9.067  & 8.636     & \underline{8.49}      & 9.139     & 9.397 & 9.136   & \textbf{8.198} \\
PIQE    & 63.816 & 64.774 & \underline{55.049} & 60.335    & 60.441    & 60.070    & 62.913 & 59.735   & \textbf{54.90} \\ \toprule[0.35mm]
\end{tabular}}
\label{tab:realrain}
\vspace{-0.4cm}
\end{table*}
\begin{table}[t!]
\centering
\caption{Comparison of quantitative results on UHD-Blur dataset~\cite{UHDformer}.}
\vspace{-10pt}
\resizebox{\linewidth}{!}{
\begin{tabular}{c|c|c|cc|c}
\toprule[0.5mm]
Methods     & Type                              & Venue     & PSNR    & SSIM  & Param  \\ \hline
MIMO-Unet++~\cite{MIMO} & \multicolumn{1}{c|}{\multirow{5}{*}{non-UHD}} & ICCV'21 & 25.03 & 0.752 & 16.1M\\
Restormer~\cite{Restormer}   & \multicolumn{1}{c|}{}                         & CVPR'22 & 25.21 & 0.752 & 26.1M\\
Uformer~\cite{Uformer}     & \multicolumn{1}{c|}{}                         & CVPR'22 & 25.27 & 0.752 & 20.6M\\
Stripformer~\cite{Stripformer} & \multicolumn{1}{c|}{}                         & ECCV'22 & 25.05 & 0.750 & 19.7M\\
FFTformer~\cite{fftformer}   & \multicolumn{1}{c|}{}                         & CVPR'23 & 25.41 & 0.757 & 16.6M\\ \hline
UHDformer~\cite{UHDformer}   & \multicolumn{1}{c|}{\multirow{6}{*}{UHD}}     & AAAI'24 & 28.82 & 0.844 & 0.34M\\
UHDDIP~\cite{UHDDIP}      & \multicolumn{1}{c|}{}                         & TCSVT'25& 28.28 & 0.845 & 0.81M\\
DreamUHD~\cite{DreamUHD}    & \multicolumn{1}{c|}{}                         & AAAI'25 & 29.33 & 0.852 & 1.45M\\
UHD-processer~\cite{UHD-processer} & \multicolumn{1}{c|}{}                       & CVPR'25 & 29.43 & 0.855 & 1.6M\\
ERR~\cite{ERR}         & \multicolumn{1}{c|}{}                         & CVPR'25  & \underline{29.72} & \underline{0.861} & 1.13M\\
Ours        & \multicolumn{1}{c|}{}                         & -       & \textbf{31.53} & \textbf{0.890} & 2.71M\\
\bottomrule[0.5mm]
\end{tabular}}
\label{tab:UHDBlur}
\vspace{-0.4cm}
\end{table}

\noindent\textbf{Image Deraining Results}: 
To validate the effectiveness of our method on the task of rain streak Removal, we compare it with many methods on 4K-Rain13k dataset~\cite{UDR-Mixer} and 4K-RealRain dataset~\cite{UDR-Mixer}, including RCDNet~\cite{RCDNet}, SPDNet~\cite{SPDNet}, IDT~\cite{IDT}, Restormer~\cite{Restormer}, DRSformer~\cite{DRSformer}, UDR-S2Former~\cite{UDR-S2Former}, NeRD-Rain~\cite{NeRD-Rain}, MambaIRv2~\cite{mambairv2}, UDR-Mixer~\cite{UDR-Mixer} and ERR~\cite{ERR}. As demonstrated in Tabs.~\ref{tab:4KRain} and~\ref{tab:realrain}, our method achieves SOTA performance consistently across both synthetic and real-world datasets. Specifically in the 4K-Rain13k dataset, it delivers significant PSNR improvements of 1.96 dB over MambaIRv2 and 0.65 dB over ERR, both Mamba-based methods. Furthermore, visual comparisons in Fig.~\ref{fig:rain} provide additional validation of our method's efficacy.

\begin{figure*}[t!]
    \centering
    \includegraphics[width=.85\linewidth]{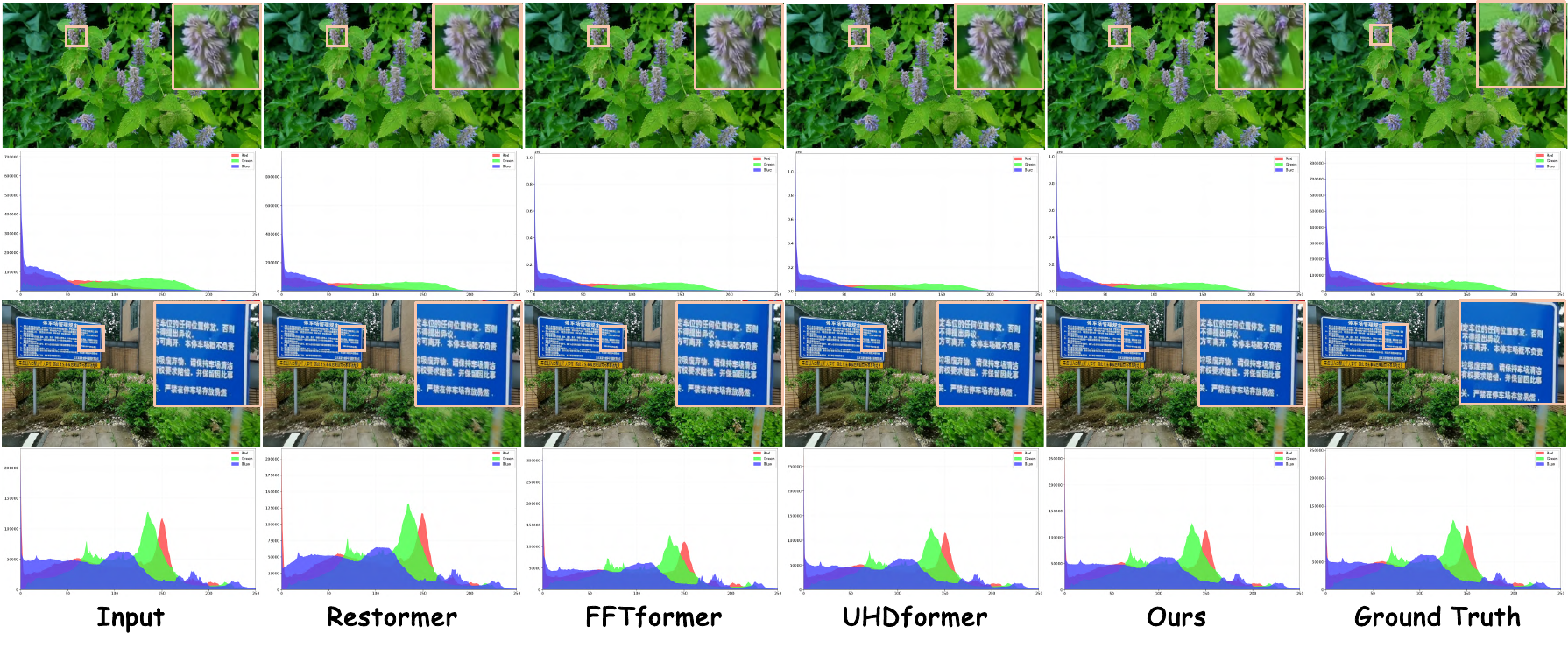}
    \vspace{-10pt}
    \caption{Visual quality comparisons on UHD-Blur dataset~\cite{UHDformer}. The last row shows the color histogram of the image.}
    \label{fig:blur}
    \vspace{-8pt}
\end{figure*}

\noindent\textbf{Image Deblurring Results}: 
In the UHD image deblurring task~\cite{UHDformer}, we evaluate our proposed C$^2$SSM against existing deblurring approaches, including MIMO-Unet++~\cite{MIMO}, Restormer~\cite{Restormer}, Uformer~\cite{Uformer}, Stripformer~\cite{Stripformer}, FFTformer~\cite{fftformer}, UHDformer~\cite{UHDformer}, UHDDIP~\cite{UHDDIP}, DreamUHD~\cite{DreamUHD}, UHD-processer~\cite{UHD-processer} and ERR~\cite{ERR}. As quantified in Tab.~\ref{tab:UHDBlur}, our method achieves a significant 1.81 dB PSNR advantage over the top-performing baseline ERR. Visual evidence in Fig.~\ref{fig:blur} corroborates that our reconstructions exhibit superior structural integrity and visual naturalness.

\noindent\textbf{Image Dehazing Results}: 
For UHD image dehazing task~\cite{UHDformer}, we compare our C$^2$SSM with a wide range of state-of-the-art methods, including Restormer~\cite{Restormer}, Uformer~\cite{Uformer}, DehazeFormer~\cite{dehazeformer}, MB-TaylorFormer~\cite{taylorformer}, UHD~\cite{UHDHDR}, UHDformer~\cite{UHDformer}, UHDDIP~\cite{UHDDIP} and UHD-processer~\cite{UHD-processer}. As shown in Tab.~\ref{tab:UHDHaze}, our method achieves favorable results in quantitative metrics compared to existing approaches.

\noindent\textbf{Image Desnowing Results}: 
In the image desnowing task, we compare HiFormer with Uformer~\cite{Uformer}, Restormer~\cite{Restormer}, SFNet~\cite{SFNet}, UHD~\cite{UHDHDR}, UHDformer~\cite{UHDformer}, and UHDDIP~\cite{UHDDIP}. As shown in Tab.~\ref{tab:UHDSnow}, our method exceeds the current best-performing method, UHDDIP, by 1.5 dB in PSNR.

\subsection{Ablation Studies and Discussions}
We further conduct extensive ablation studies to better understand and evaluate each component in the proposed C$^2$SSM. For a fair comparison, all these variants are trained using the same settings. 

\begin{table}[t!]
\centering
\caption{Comparison of quantitative results on UHD-Haze dataset~\cite{UHDformer}.}
\vspace{-10pt}
\resizebox{\linewidth}{!}{
\begin{tabular}{c|c|c|cc|c}
\toprule[0.5mm]
Methods         & Type                                          & Venue   & PSNR  & SSIM  & Param \\ \hline
Restormer~\cite{Restormer}       & \multicolumn{1}{c|}{\multirow{4}{*}{non-UHD}} & CVPR'22 & 12.72 & 0.693 & 26.11M\\
Uformer~\cite{Uformer}         & \multicolumn{1}{c|}{}                         & CVPR'22 & 19.83 & 0.737 & 20.63M\\
DehazeFormer~\cite{dehazeformer}    & \multicolumn{1}{c|}{}                         & TIP'23  & 15.37 & 0.725 & 2.5M\\
MB-TaylorFormer~\cite{taylorformer} & \multicolumn{1}{c|}{}                         & ICCV'23 & 20.99 & 0.919 & 2.7M\\\hline
UHD~\cite{UHDHDR}             & \multicolumn{1}{c|}{\multirow{5}{*}{UHD}}     & ICCV'21 & 18.04 & 0.811 & 34.5M\\
UHDformer~\cite{UHDformer}       & \multicolumn{1}{c|}{}                         & AAAI'24 & 22.59 & \underline{0.942} & 0.34M \\
UHDDIP~\cite{UHDDIP}          & \multicolumn{1}{c|}{}                         & TCSVT'25& 22.14 & 0.941 &  0.81M\\
UHD-processer~\cite{UHD-processer}   & \multicolumn{1}{c|}{}                         & CVPR'25 & \underline{23.24} & \textbf{0.953} &  1.6M\\
Ours            & \multicolumn{1}{c|}{}                         & -       & \textbf{24.08} & \underline{0.942} & 2.71M\\
\bottomrule[0.5mm]
\end{tabular}}
\label{tab:UHDHaze}
\vspace{-0.4cm}
\end{table}
\begin{table}[t!]
\centering
\caption{Comparison of quantitative results on UHD-Snow dataset~\cite{UHDDIP}.}
\vspace{-10pt}
\resizebox{\linewidth}{!}{
\begin{tabular}{c|c|c|cc|c}
\toprule[0.5mm]
Methods   & Type                     & Venue   & PSNR  & SSIM  & Param  \\ \hline
Uformer~\cite{Uformer}   & \multirow{3}{*}{non-UHD} & CVPR'22 & 23.71 & 0.871 & 20.63M \\
Restormer~\cite{Restormer} &                          & CVPR'22 & 24.14 & 0.869 & 26.12M \\
SFNet~\cite{SFNet}     &                          & ICLR'23 & 23.63 & 0.845 & 7.05M  \\ \hline
UHD~\cite{UHDHDR}       & \multirow{4}{*}{UHD}     & ICCV'21 & 29.29 & 0.949 & 34.5M  \\
UHDformer~\cite{UHDformer} &                          & AAAI'24 & 36.61 & \underline{0.988} & 0.34M  \\
UHDDIP~\cite{UHDDIP}    &                          & TCSVT'25& \underline{41.56} & \textbf{0.990} & 0.81M  \\
Ours      &                          & -       & \textbf{42.45} & \textbf{0.990}    & 2.71M  \\ \bottomrule[0.5mm]
\end{tabular}}
\label{tab:UHDSnow}
\vspace{-0.4cm}
\end{table}

\noindent\textbf{Effectiveness of Proposed Blocks}: 
To assess the contribution of the proposed CCSM and SCFM to overall framework performance, we design a series of ablative variants involving their systematic removal or functional substitution, thereby rigorously quantifying their efficacy. As shown in Tab.~\ref{tab:block}, removing or replacing the proposed modules leads to performance degradation, confirming their effectiveness. Notably, due to computational complexity constraints, both vanilla Mamba and ASSM  are incapable of full-resolution inference on consumer-grade hardware, necessitating aggressive downsampling (typically 8$\times$) for UHD image preprocessing, followed by upsampling to restore the original resolution. This compulsory resolution compromise not only introduces additional information loss but also limits the model's capacity to preserve high-frequency details. In contrast, our CCSM, through its sparse representation mechanism based on cluster centroids, successfully achieves full-resolution processing while maintaining feasible computational overhead, fundamentally addressing the memory bottleneck in UHD image restoration. The most significant performance drop occurs when CCSM is removed, demonstrating that long-range dependency modeling is crucial for the restoration task. Furthermore, CCSM proves superior to SCFM in importance, as SCFM essentially serves as a supplementary module to compensate for information loss caused by incomplete global modeling in CCSM.

\begin{table}[t]
\centering
\caption{Ablation study of proposed blocks on UHD-LOL4K dataset~\cite{LLFormer}.}
\vspace{-10pt}
\resizebox{\linewidth}{!}{
\begin{tabular}{l|ccc}
\toprule[0.5mm]
\multicolumn{1}{c|}{Variants}  & PSNR  & SSIM  & Param \\ \hline
(a) replace CCSM with ResBlock & 37.32 & 0.989 & 2.38M \\
(b) replace CCSM with Vanilla Mamba~\cite{mambair} & 37.43 & 0.990 & 3.11M \\
(c) replace CCSM with ASSM~\cite{mambairv2} & 38.71 & 0.991 & 2.96M \\
(d) replace SCFM with ResBlock & 39.25 & 0.992 & 2.69M \\
(e) remove CCSM                & 35.87 & 0.987 & \textbf{2.21M} \\
(f) remove SCFM                & 39.05 & 0.992 & 2.53M \\ \hline
(g) full model (ours)          & \textbf{39.61} & \textbf{0.992} & 2.71M \\
\toprule[0.5mm]
\end{tabular}}
\label{tab:block}
\vspace{-0.4cm}
\end{table}

\noindent\textbf{Validation of the Number of Centers}: 
To investigate the impact of cluster center quantity on model performance, we conduct experiments on UHD-LOL4K, UHD-Blur, and UHD-Haze datasets. Results in Tab.~\ref{tab:center} demonstrate that setting the number of centers to 4 achieves optimal balanced performance across multiple datasets. As the center count increases, average performance slightly declines, potentially due to redundant clusters affecting results. Notably, the UHD-Blur dataset attains peak performance with 6 centers, which may be attributed to its inherent dataset bias: consisting predominantly of indoor and complex scenes with high content diversity, demanding more complex representations. In contrast, UHD-LOL4K and UHD-Haze primarily contain outdoor scenes featuring extensive homogeneous regions such as skies and ground planes.

\noindent\textbf{Comparison with Other Scanning Strategy}: 
To validate the efficiency of our proposed scanning strategy, we compare computational complexity with various Mamba-based methods. While MambaIR~\cite{mambair} and Wave-Mamba~\cite{wavemamba} employ vanilla scanning strategies, both EVSSM~\cite{evssm} and MambaIRv2~\cite{mambairv2} implement customized scanning designs. As demonstrated in Tab.~\ref{tab:diff}, our approach achieves the lowest computational complexity. It is worth noting that, except for Wave-Mamba, these methods cannot perform full-resolution inference on UHD images, so we do not report their performance uniformly. You can find the comparison results of Wave-Mamba and MambaIRv2 with our method in Tab.~\ref{tab:UHDLOL},~\ref{tab:UHDLL},~\ref{tab:4KRain} and~\ref{tab:realrain}. Collectively, the CCSM design enables our method to maintain outstanding performance while significantly reducing computational overhead.

\begin{table}[t!]
\centering
\caption{Ablation study of the number of centers.}
\vspace{-10pt}
\resizebox{\linewidth}{!}{
\begin{tabular}{c|cccc}
\toprule[0.5mm]
Number of Centers & 2           & 4           & 6           & 8           \\ \hline
UHD-LOL4K~\cite{LLFormer}         & 37.91/0.992 & 39.61/0.992 & 39.23/0.992 & 39.19/0.992 \\
UHD-Blur~\cite{UHDformer}          & 30.42/0.881 & 31.53/0.890 & 31.72/0.890 & 31.65/0.889 \\ 
UHD-Haze~\cite{UHDformer}          & 23.27/0.940 & 24.08/0.942 & 24.03/0.942 & 23.86/0.942 \\ 
\toprule[0.5mm]
\end{tabular}}
\label{tab:center}
\vspace{-0.3cm}
\end{table}
\begin{table}[t]
\centering
\caption{Comparison of different scanning strategies. FLOPs are measured with an image of the size $64\times64$ pixels.}
\vspace{-10pt}
\resizebox{\linewidth}{!}{
\begin{tabular}{c|ccccc}
\toprule[0.5mm]
Method & MambaIR~\cite{mambair} & Wave-Mamba~\cite{wavemamba} & EVSSM~\cite{evssm}   & MambaIRv2~\cite{mambairv2} & Ours   \\ \hline
Parm   & 1.32M   & 1.25M      & 17.13M  & \textbf{0.84M}     & 2.71M  \\
FLOPs  & 4.774G  & 0.881G     & 7.893G  & 4.940G    & \textbf{0.407G} \\
Venue  & ECCV'24 & MM'24      & CVPR'25 & CVPR'25   & -      \\ 
\toprule[0.5mm]
\end{tabular}}
\label{tab:diff}
\vspace{-0.4cm}
\end{table}
\section{Conclusion}
In this paper, we proposed C$^2$SSM, a novel visual state space model that breaks the computational bottlenecks of existing mamba-based methods in UHD image restoration by shifting from pixel-serial to cluster-serial scanning. The core of C$^2$SSM lies in the CCSM, which models UHD images as a sparse set of semantic centroids. By performing global reasoning only on these centroids and diffusing the learned context back to pixels via a principled similarity distribution, CCSM achieves a dramatic reduction in computational complexity without sacrificing performance. Complementing this, the SCFM ensures the preservation of high-frequency details that may be overlooked during clustering. CCSM and SCFM complement each other. Comprehensive experiments across numerous UHD image restoration tasks reveal our method surpasses current SOTA methods in both quantitative metrics and qualitative analysis.



{
    \small
    \bibliographystyle{ieeenat_fullname}
    \bibliography{main}
}


\end{document}